\documentclass[runningheads]{llncs}
\usepackage[T1]{fontenc}
\usepackage{graphicx,verbatim}
\usepackage{hyperref}

\usepackage{color}

\usepackage{amsmath}
\usepackage{tikz}
\usepackage{graphicx}
\usepackage{multirow}
\usepackage{booktabs}
\usepackage{tabularx}
\usepackage{array}
\usepackage{adjustbox}
\usepackage{placeins}

\begin{document}
%
\title{Mamba-driven MRI-to-CT Synthesis for MRI-only Radiotherapy Planning}

\titlerunning{Mamba-driven MRI-to-CT Synthesis}
%

\author{Konstantinos Barmpounakis\inst{1,2} 
\and {Theodoros P. Vagenas}\inst{1} 
\and Maria Vakalopoulou\inst{2,3} 
\and George K. Matsopoulos\inst{1}}   

\authorrunning{K. Barmpounakis et al.}
\institute{School of Electrical and Computer Engineering, National Technical University of Athens \\
\email{kbarbounakis@biomed.ntua.gr} 
    \and Archimedes, Athena Research Center 
    \and Cancer Data Science Unit, Université Paris-Saclay, Gustave Roussy, CentraleSupélec, INSERM, Villejuif, France. 
    }
  
\maketitle              
\begin{abstract}
Radiotherapy workflows for oncological patients increasingly rely on multi-modal medical imaging, commonly involving both Magnetic Resonance Imaging (MRI) and Computed Tomography (CT). MRI-only treatment planning has emerged as an attractive alternative, as it reduces patient exposure to ionizing radiation and avoids errors introduced by inter-modality registration. While nnU-Net-based frameworks are predominantly used for MRI-to-CT synthesis, we explore Mamba-based architectures for this task to investigate the applicability of state-space modeling for cross-modality medical image translation and assess its performance relative to established convolutional architectures. Specifically, we adapt the SegMamba architecture, originally proposed for segmentation, to perform image-to-image generation. Our 3D Mamba architecture effectively captures complex volumetric features and long-range dependencies, thus allowing accurate CT synthesis while maintaining a relatively low parameter count. Experiments were conducted on a subset of SynthRAD2025 dataset, comprising registered single-channel MRI–CT volume pairs across three anatomical regions. Quantitative evaluation is performed via a combination of image similarity metrics computed in Hounsfield Units (HU) and segmentation-based metrics obtained from TotalSegmentator to ensure geometric consistency is preserved. The findings pave the way for the integration of state-space models into radiotherapy workflows.


\end{abstract}
\section{Introduction}
Accurate imaging forms the foundation of diagnoses and treatment protocols in oncology, enabling effective clinical decisions. In the context of radiotherapy, the complementary strengths of Magnetic Resonance Imaging (MRI) and Computed Tomography (CT) are leveraged to attain precise tumor targeting with optimal treatment outcomes while minimizing exposure to healthy tissues~\cite{MRI_radiotherapy}. The former modality allows for superior soft-tissue contrast and tumor outlining accuracy, whereas the latter contributes to reliable dose calculation due to the inherent relationship between Hounsfield Units (HU) and electron density. 

Since radiotherapy is typically delivered in weekday fractions over several weeks, it would ideally require MRI and CT scans prior to each treatment session. However, repeated CT acquisition increases patient exposure to ionizing radiation beyond the therapeutic dose and is not always technically feasible or financially viable in certain clinical settings. Conventional workflows require registration approaches in order to transfer the anatomical delineations from MRI onto the CT scan used for dose calculation. The accumulation of registration errors may further degrade the treatment performance. Therefore, a simplified MRI-only workflow emerges as a promising alternative, eliminating the need for multiple CT scans~\cite{sCT_review,MRI_only_planning}. Recent advances in medical imaging computation, including cross-modality translation using deep learning techniques, enable the accurate synthesis of CT images directly from MRI data~\cite{deepLearning_sCT}.

A major limitation in the development of MRI-only workflows is the scarcity of large-scale, heterogeneous and well-curated paired MRI-CT datasets. Constructing such datasets is a tedious and resource-intensive process, encompassing multi-institutional coordination, standardized acquisition protocols, quality control, ethical approval as well as carefully designed pre-processing pipelines and registration procedures. Despite these constraints, several studies have explored MRI-only workflows leveraging unpaired data in an unsupervised or semi-supervised manner. For instance, Brou Boni et al. proposed an unpaired CycleGAN framework to synthesize CT from MRI eliminating the need for paired data~\cite{unpairedMRI-to-CT_CycleGAN}, while MaskGAN enforces anatomical consistency through automatically extracted structural masks during unpaired MR–CT translation~\cite{unpairedMRI-to-CT_maskgan}.

This paper introduces the use of Mamba-based architectures for MRI-to-CT synthesis across multiple anatomical regions. To the best of our knowledge, this work represents the first integration of the SegMamba framework into a generative medical imaging task~\cite{segmamba}. We further propose a compound loss scheme specifically designed for this task and ultimately demonstrate the robustness and the efficiency of the Mamba modules in CT synthesis, comparing their performance with the conventional nnU-Net framework.

\section{Methodology}

\subsubsection{Architectural Design} Convolutional Neural Networks (CNNs) have long been the dominant paradigm for 3D medical image tasks due to their strong inductive bias toward local spatial context, enabling precise voxel-wise predictions and stable training even with limited data. Building upon the standard U-Net design, nnU-Net introduces a self-configuring framework that automatically adapts preprocessing, normalization, network topology and training settings to a given dataset, providing a robust baseline by eliminating manual hyper-parameter tuning~\cite{nnunet,nnunetv2}. Despite their effectiveness, purely convolutional architectures exhibit limited receptive fields and struggle to capture long-range dependencies inherent in diverse applications. 

State-space models (SSMs) perform sequential data processing through latent dynamical systems combining long-range modeling capabilities with favorable computational scaling~\cite{ssm}. Mamba introduces selective state-space blocks that integrate SSMs productively into deep learning frameworks. Motivated by the complementary strengths of CNNs and SSMs, we propose FlexibleMamba, a SegMamba-inspired framework for volumetric MRI-to-CT synthesis. Building upon the dual-encoder design introduced in SegMamba, FlexibleMamba retains the hierarchical MambaEncoder for global context modeling together with a UNETR encoder-decoder reconstruction pathway, while introducing targeted architectural modifications that better suit continuous image synthesis~\cite{segmamba,mamba,unetr}. 

\subsubsection{Mamba Encoder} The primary feature extraction pipeline relies on the MambaEncoder, operating across five hierarchical stages. At each stage, spatial resolution is halved and channel capacity is amplified via a Downsampling module comprising Layer Normalization followed by a Convolutional Block. 

Following downsampling, the extracted features are processed by an FMamba block, which consists of a Gated Spatial Convolution (GSC), two sequential Mamba layers, and a channel-wise Multi-Layer Perceptron (MLP). The GSC first enhances local spatial representations using residual convolutions. In contrast to the original SegMamba implementation, we replace the ReLU activations with Leaky ReLU , promoting more stable gradient propagation and eliminating blurriness in the output. The refined features are subsequently processed by two consecutive Mamba layers. To efficiently model long-range spatial dependencies, each layer temporarily reshapes the volumetric feature map into a sequence of tokens, applies Layer Normalization and state-space modeling and finally restores the original volumetric arrangement before residual aggregation. A channel-wise MLP then performs feature refinement through an expansion–projection scheme ($C\rightarrow2C\rightarrow C$), producing the output representation for the next hierarchical stage.

\subsubsection{UNETR Encoder} To bridge the semantic gap between deep sequential features and the spatial reconstruction task, outputs from the MambaEncoder are injected into the localized ResBlocks of the parallel UNETR Encoder, which subsequently provide multi-scale skip connections to the decoder.

\subsubsection{UNETR Decoder}Image reconstruction is performed by a UNETR-style decoder comprising five successive UpBlocks. Each UpBlock upsamples the incoming feature map using a transposed convolution, fuses it with the corresponding encoder skip connection and refines the fused representation through a residual block. After the final upsampling stage, an additional residual block further refines the highest-resolution representation before a conventional convolutional block projects the features to form the synthesized CT patch.

Unlike the original SegMamba architecture, all residual blocks in both the UNETR encoder and decoder employ Group Normalization instead of Instance Normalization. While Instance Normalization is well suited to segmentation, its normalization of each channel independently tends to suppress absolute intensity statistics that are essential for image synthesis. Group Normalization preserves richer inter-channel statistics while remaining robust under the small batch sizes commonly encountered in 3D volumetric learning, leading to more precise MRI-to-CT intensity translation.

\begin{figure}
\includegraphics[width=\textwidth]{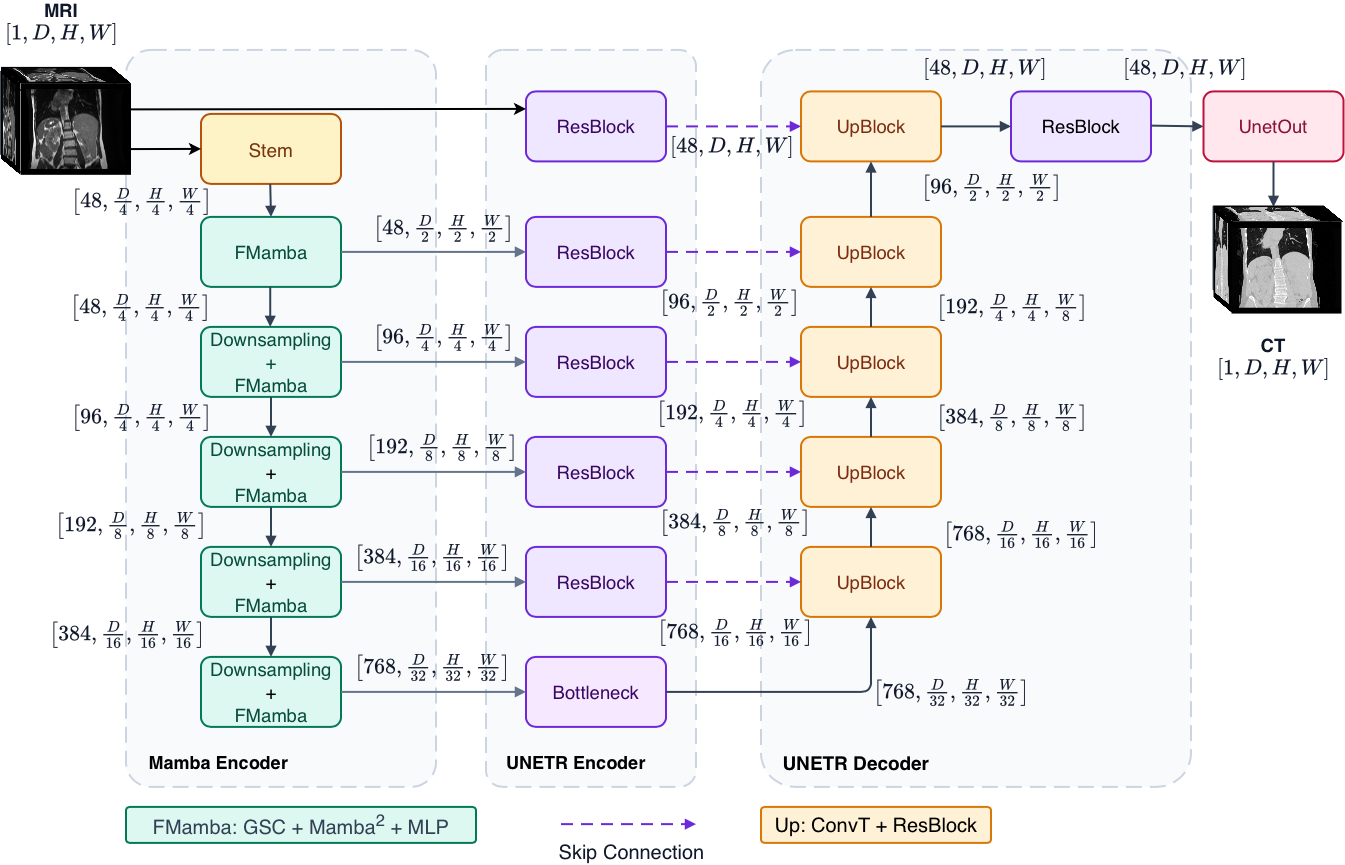}
\caption{\textbf{Overview of the FlexibleMamba architecture for 3D MRI-to-CT synthesis.} The input MRI patch is processed by the MambaEncoder and parallel UNETR encoder, generating hierarchical features that are refined and propagated through skip connections to the decoder. FMamba blocks combine gated spatial convolutions, sequential Mamba layers, and channel-wise feature mixing. Successive UpBlocks restore spatial resolution and produce the final synthetic CT.}
\label{fig:flexible_mamba}
\label{overview}
\end{figure}

\subsubsection{Loss function} To ensure the synthetic CT volumes exhibit both accurate voxel intensities and high structural fidelity, the network is optimized using a compound loss function. The overall training objective has the following structure:
\begin{equation}
    \mathcal{L} = \mathcal{L}_{MAE} + w_{1}\mathcal{L}_{AFP} + w_{2}\mathcal{L}_{MS-SSIM}
\label{eq:loss}    
\end{equation}

Each component addresses some specific requirement, with $w_{1}$ and $w_{2}$ acting as empirically selected weighting factors to balance the terms:

\begin{itemize}
    \item \textbf{Mean Absolute Error (MAE):} The $\mathcal{L}_{MAE}$ term serves as the foundational pixel-level regression loss. It directly minimizes the absolute HU deviations between the generated CT and the ground truth CT, optimizing the model for global intensity accuracy.
    \item \textbf{Anatomical Feature-Prioritized (AFP) Loss:} Arthur et al. recently introduced the Anatomical Feature-Prioritized (AFP) loss, a novel objective function designed to enhance the reconstruction of localized anatomical details in medical image synthesis~\cite{afp}. Unlike conventional pixel-wise losses that primarily optimize global intensity similarity, AFP operates in feature space by comparing multi-scale embeddings extracted from a pre-trained foundational model. Inspired by the principles of perceptual loss, AFP exploits hierarchical features optimized for anatomical delineation, encouraging the conformation of clinically relevant structures rather than solely minimizing voxel-wise differences. In our framework, the computation of the $\mathcal{L}_{AFP}$ term relies on multi-scale embeddings generated by TotalSegmentator~\cite{totalsegmentator}, guiding the synthesis process toward improved geometric consistency and anatomically meaningful reconstructions.
    \item \textbf{Multi-Scale Structural Similarity Index Measure (MS-SSIM):} The $\mathcal{L}_{MS-SSIM}$ term, defined as $\mathcal{L}_{\text{MS-SSIM}} = 1 - \text{MS-SSIM}$, encourages structural consistency between the synthesized and reference CT volumes. Unlike single-scale SSIM, MS-SSIM evaluates structural similarity over progressively downsampled image representations, capturing both fine anatomical details and coarse structural context.
\end{itemize}

\subsubsection{Quantitative Evaluation} Quantitative evaluation is performed using standard image similarity metrics, including MAE, Peak Signal-to-Noise Ratio (PSNR) and Multi-Scale SSIM (MS-SSIM), which extends SSIM by averaging structural similarity across multiple resolution scales~\cite{image_similarity_metrics}. In addition, anatomical consistency is assessed through downstream segmentation analysis by computing Dice Similarity Coefficient (DSC) and the $95^{th}$ percentile of Hausdorff Distance (HD95) between several masks obtained with TotalSegmentator on the generated and target CT volumes~\cite{totalsegmentator,HD95}.

\section{Experiments}

\subsection{Dataset \& Preprocessing}

The SynthRAD2023 Grand Challenge dataset was developed to address the scarcity of co-registered multimodal data, providing 540 rigidly registered MRI–CT pairs from three Dutch medical centers across the brain and pelvis, alongside paired cone-beam computed tomography (CBCT)–CT data for image synthesis~\cite{synthRAD2023}. Building on this foundation, the SynthRAD2025 Grand Challenge dataset substantially expands both the scale and diversity of paired data, spanning head‑and‑neck, thoracic and abdominal sites across five European medical centers~\cite{synthRAD2025}.

The current work utilizes a curated subset of the SynthRAD2025 dataset. In particular, we focus on the MRI–CT synthesis subtask, which comprises paired data from four European medical centers across three anatomical regions: abdomen (AB), head-and-neck (HN), and thorax (TH). Data from center D were excluded due to usage restrictions. The MRI sequence used varied according to the body region and the acquisition protocol of each medical center, consisting of either T1-weighted gradient-echo or balanced steady-state free-precession sequences. In total, 890 MRI–CT pairs are provided together with an outline mask defining the patient's body region. The dataset follows an official 65\%/10\%/25\% split into training, validation and test sets, respectively, where only the training percentage is fully accessible. Consequently, we used the training set and further divided it into 90\%/10\% subsets, employing the larger portion for model development and keeping the remaining 10\% fully unseen until the evaluation procedure.

To ensure the highest quality of training data and prevent the model from learning erroneous cross-modality mappings, a rigorous visual inspection and data curation step was performed. In total, 70 samples were excluded, including 30 abdomen, 22 thorax and 18 head-and-neck cases, owing to substantial inter-modality misregistration or incomplete field-of-view (FOV) coverage in either modality. Consequently, the resulting subset encompasses 396 training cases and 45 test cases, distributed across the three anatomical regions as follows: 129 training and 15 test cases for AB, 124 and 13 for HN, and 143 and 17 for TH, respectively.

The provided SynthRAD2025 data have undergone an initial preprocessing pipeline including rigid registration of MRI to CT using the Elastix framework to ensure spatial alignment~\cite{elastix,elastix2}. Facial structures were removed for anonymization and then all volumes were resampled to a common voxel spacing of $3\times1\times1$\,mm before being cropped with a 10-pixel margin around automatically generated patient outline masks. In addition, we apply supplementary preprocessing performing per-scan z-score normalization of MRI volumes and subsequent clipping to $[-5,5]$ to suppress extreme outliers. CT volumes were clipped to $[-1024,3000]$ HU prior to scaling by a factor of $1000$.

\subsection{Implementation Details}

All models were trained from scratch on an NVIDIA A100-SXM4-80GB GPU using a patch size of $[64,224,224]$ and a batch size of 2, without additional data augmentation. Default self-configuration procedures were retained for nnU-Net, adapted to the generative setting utilizing \emph{nnUNet\_translation} repository~\cite{nnunet_translation}. Optimization was performed using AdamW with an initial learning rate of $3\times10^{-4}$ and a polynomial decay schedule. All models were trained for 1500 epochs except nnU-Net which was trained for 1000 epochs following its default training schedule. The proposed compound loss function was used consistently across all experiments, where the weighting coefficients in Eq.~\eqref{eq:loss} were empirically determined through a small-scale hyperparameter search conducted in the training data and set to $w_1=0.02$ and $w_2=0.3$. The source code and implementation details are publicly available at \url{https://github.com/kbarbou/mamba-driven-mri2ct}.

\begin{figure*}[!t]
\includegraphics[width=\textwidth]{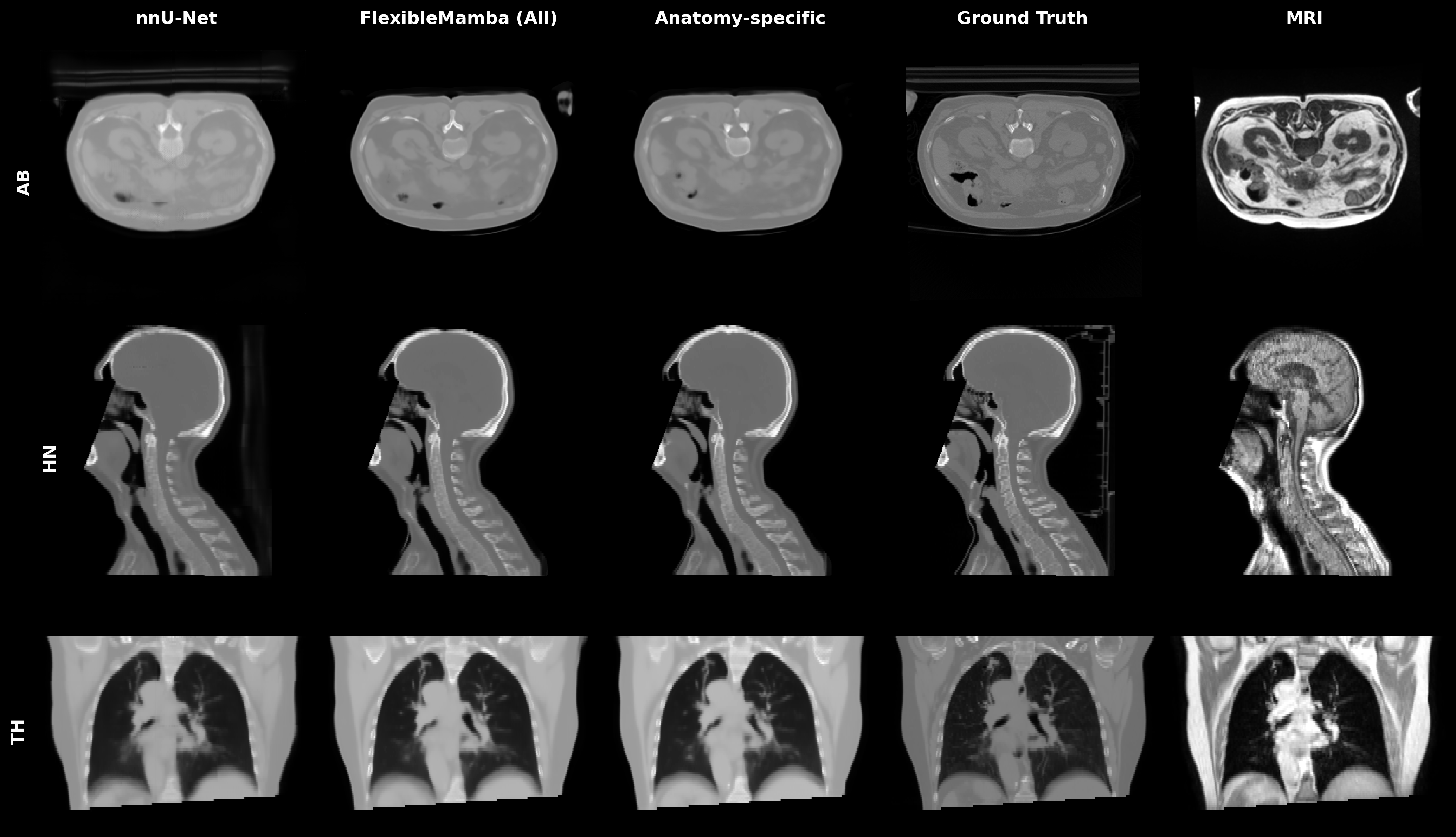}
\caption{Visual comparison in CT synthesis across the three anatomical regions. Columns (left to right) depict synthetic CT volumes generated by nnU-Net, the proposed FlexibleMamba trained jointly on all regions (FlexibleMamba-All), anatomy-specific FlexibleMamba models, the corresponding ground-truth CT, and the input MRI.} \label{qualitative}

\end{figure*}

\subsection{Results}
The quantitative comparison is summarized in Table~\ref{tab:quantitative_results}, while representative qualitative examples are presented in Figure~\ref{qualitative}. All voxel-wise image quality metrics were computed exclusively within the provided outline masks, such that only voxels belonging to the masked region contributed to the evaluation. Across all anatomical regions, nnU-Net achieves the best overall quantitative performance as well as slightly improved geometric consistency. This result is expected given nnU-Net's mature self-configuring optimization pipeline and its strong convolutional inductive bias for dense medical image reconstruction. Nevertheless, region-wise statistical analysis reveals that no statistically significant differences between FlexibleMamba and nnU-Net are observed for MAE, DSC, or HD95 in either the head-and-neck ($p=0.092$, $0.126$ and $0.393$ respectively) or thoracic ($p=0.092$, $0.126$ and $0.070$ respectively) regions.  Furthermore, the qualitative comparison demonstrates that FlexibleMamba consistently generates plausible synthetic CT volumes while preserving major bone structures as well as soft-tissue interfaces.

In terms of the unified FlexibleMamba and the anatomy-specific variants trained independently for the abdomen, head-and-neck, and thorax regions, the statistical analysis provides no evidence that region-specific training offers a systematic advantage. On the contrary, the unified FlexibleMamba significantly improves DSC and HD95 in the abdomen, MAE and PSNR in the head-and-neck region and all five evaluated metrics in the thoracic region compared with the corresponding anatomy-specific models. These findings suggest that the unified FlexibleMamba effectively learns anatomy-invariant representations capable of generalizing across heterogeneous anatomical regions without requiring separate region-specific models.

Table~\ref{tab:loss} evaluates the contribution of each component of the proposed compound objective. In fact, optimizing solely with the MAE loss yields inferior structural and geometric consistency. Introducing the AFP term produces only negligible changes in the image-similarity metrics, while significantly improving DSC ($p=0.0003$) and reducing HD95 ($p<0.0001$) over the complete test cohort. This indicates that AFP primarily promotes anatomically consistent reconstructions and more accurate boundary localization without materially affecting voxel-wise fidelity. Finally, the insertion of the $\mathcal{L}{MS-SSIM}$ term results in only small numerical variations in the image-similarity metrics and maintains the improved geometric consistency of the compound objective. Overall, the results suggest that complementary supervision objectives can improve structural consistency without materially compromising voxel-wise intensity fidelity.

\begin{table*}[!t]
\centering
\caption{Quantitative comparison among nnU-Net, unified FlexibleMamba across
the abdomen (AB), head-and-neck (HN), and thorax (TH) regions and anatomy-specific variants of the FlexibleMamba. Results are
reported as mean $\pm$ standard deviation. Arrows indicate whether higher
($\uparrow$) or lower ($\downarrow$) values are preferred.}
\label{tab:quantitative_results}

\footnotesize
\setlength{\tabcolsep}{2.8pt}
\renewcommand{\arraystretch}{1.12}

\begin{adjustbox}{max width=\textwidth}
\begin{tabular}{
    @{}
    ll
    *{5}{r@{\,\(\pm\)\,}l}
    @{}
}
\toprule
\textbf{Region}
& \textbf{Model}
& \multicolumn{2}{c}{\textbf{MAE $\downarrow$}}
& \multicolumn{2}{c}{\textbf{PSNR $\uparrow$}}
& \multicolumn{2}{c}{\textbf{MS-SSIM $\uparrow$}}
& \multicolumn{2}{c}{\textbf{DSC $\uparrow$}}
& \multicolumn{2}{c}{\textbf{HD95 $\downarrow$}}
\\
\midrule

\multirow{3}{*}{\textbf{AB}}
& nnU-Net
& 98.76 & 11.97
& 25.92 & 0.83
& 0.815 & 0.038
& 0.562 & 0.123
& 11.28 & 3.65
\\

& \textbf{FlexibleMamba (All)}
& 105.51 & 15.39
& 25.13 & 1.02
& 0.791 & 0.040
& 0.519 & 0.130
& 14.22 & 5.22
\\

& FlexibleMamba (AB)
& 105.95 & 15.41
& 25.93 & 0.96
& 0.790 & 0.041
& 0.489 & 0.120
& 17.12 & 5.96
\\
\midrule

\multirow{3}{*}{\textbf{HN}}
& nnU-Net
& 95.79 & 15.58
& 26.71 & 1.48
& 0.919 & 0.037
& 0.761 & 0.078
& 4.40 & 1.18
\\

& \textbf{FlexibleMamba (All)}
& 97.60 & 17.60
& 26.40 & 1.43
& 0.915 & 0.038
& 0.745 & 0.076
& 4.93 & 1.84
\\

& FlexibleMamba (HN)
& 99.04 & 18.55
& 26.36 & 1.46
& 0.914 & 0.040
& 0.747 & 0.079
& 4.84 & 2.05
\\
\midrule

\multirow{3}{*}{\textbf{TH}}
& nnU-Net
& 94.64 & 11.85
& 26.64 & 1.04
& 0.873 & 0.027
& 0.635 & 0.072
& 7.78 & 1.46
\\

& \textbf{FlexibleMamba (All)}
& 97.91 & 12.79
& 26.02 & 0.99
& 0.858 & 0.026
& 0.607 & 0.063
& 9.41 & 2.74
\\

& FlexibleMamba (TH)
& 101.69 & 15.88
& 25.79 & 1.14
& 0.849 & 0.031
& 0.571 & 0.078
& 11.12 & 3.51
\\

\bottomrule
\end{tabular}
\end{adjustbox}
\end{table*}

\begin{table}[!t]
\centering
\caption{Ablation study of the proposed compound loss across all anatomical regions.}
\label{tab:loss}

\scriptsize 
\setlength{\tabcolsep}{2pt}
\renewcommand{\arraystretch}{1.1}

\begin{adjustbox}{max width=\linewidth}
\begin{tabular}{
    @{}
    l
    @{\hspace{5pt}}
    *{5}{r@{\,\(\pm\)\,}l}
    @{}
}
\toprule
Loss
& \multicolumn{2}{c}{MAE$\downarrow$}
& \multicolumn{2}{c}{PSNR$\uparrow$}
& \multicolumn{2}{c}{MS-SSIM$\uparrow$}
& \multicolumn{2}{c}{DSC$\uparrow$}
& \multicolumn{2}{c}{HD95$\downarrow$}
\\
\midrule

MAE
& 99.95 & 16.17
& 25.79 & 1.29
& 0.850 & 0.061
& 0.586 & 0.143
& 14.02 & 8.65
\\

MAE + AFP
& 100.02 & 16.08
& 25.81 & 1.27
& 0.852 & 0.060
& 0.614 & 0.121
& 10.02 & 5.00
\\

MAE + AFP + MS-SSIM
& 100.35 & 15.60
& 25.83 & 1.26
& 0.852 & 0.060
& 0.617 & 0.013
& 9.72 & 5.13
\\

\bottomrule
\end{tabular}
\end{adjustbox}
\end{table}

\section{Conclusion}
The current study probes the suitability of Mamba-based architectures for MRI-to-CT synthesis within the context of radiotherapy planning. Experiments were conducted on a curated subset of the SynthRAD2025 dataset across three anatomical regions, where FlexibleMamba extends the SegMamba framework to the image-to-image translation task. In addition, we explored a compound loss that combines complementary supervision objectives to jointly promote voxel-wise intensity fidelity and structural consistency. Overall, the results demonstrate the potential of selective state-space models for cross-modality translation. The absence of dosimetric validation, due to the lack of reference treatment plans and dose distributions in the SynthRAD2025 training cohort, limits conclusions regarding the clinical suitability of the generated synthetic CTs for radiotherapy dose calculation.

\FloatBarrier
\begin{credits}
\subsubsection{\ackname} The authors would like to thank the organizers of the SynthRAD2025 Challenge for providing this multi-centric dataset and the comprehensive evaluation framework that made this research possible. Also, this work has been partially supported by project MIS 5154714 of the National Recovery and Resilience Plan Greece 2.0 funded by the European Union under the NextGenerationEU Program. We gratefully acknowledge NVIDIA Corporation for the GPU hardware grant that facilitated the conducted computational experiments.

\subsubsection{\discintname}
The authors declare that they have no competing interests.
\end{credits}

%
%
%
%
\FloatBarrier

\end{document}